\documentclass[letterpaper, 10 pt, conference]{ieeeconf}  % Comment this line out if you need a4paper
\IEEEoverridecommandlockouts                              % This command is only needed if
\usepackage{graphicx} % for pdf, bitmapped graphics files
\usepackage{times} % assumes new font selection scheme installed
\usepackage{cite}
\usepackage{amsmath} % assumes amsmath package installed
\usepackage{amssymb} % assumes amsmath package installed
\usepackage{amsfonts} % blackboard math symbols
\usepackage{bbm}
\usepackage{braket}
\usepackage{booktabs}
\usepackage{colortbl}
\usepackage{url}
\usepackage{mathtools}
\usepackage{dashrule}

\usepackage[linesnumbered,ruled,vlined]{algorithm2e}
\SetKwInput{KwInput}{Input}                % Set the Input
\SetKwInput{KwOutput}{Output}              % set the Output

\usepackage[caption=false, font=footnotesize]{subfig}
\usepackage{lipsum}
\usepackage{widetext}

\title{\LARGE \textbf{Domain-Adversarial and -Conditional State Space Model \\ for Imitation Learning}}

\author{Ryo Okumura$^{*,\star}$, Masashi Okada$^{\dag}$ and Tadahiro Taniguchi$^{\dag,\ddag}$ % <-this % stops a space
  \thanks{$^{*}$ Ryo Okumura is with Core Element Technology Development Center, Panasonic Corporation, Japan.}%
  \thanks{$^{\dag}$ Masashi Okada and Tadahiro Taniguchi are with AI Solutions Center, Business Innovation Division, Panasonic Corporation, Japan.}%
  \thanks{$^{\ddag}$ Tadahiro Taniguchi is also with Ritsumeikan University, College of Information Science and Engineering, Japan.}%
  \thanks{$^{\star}$ Corresponding author: \texttt{okumura.ryo001@jp.panasonic.com}}
  }

\begin{document}

\maketitle
\thispagestyle{empty}
\pagestyle{empty}

%%%%%%%%%%%%%%%%%%%%%%%%%%%%%%%%%%%%%%%%%%%%%%%%%%%%%%%%%%%%%%%%%%%%%%%%%%%%%%%%
\begin{abstract}
  State representation learning (SRL) in partially observable Markov decision processes has been studied to learn abstract features of data useful for robot control tasks. For SRL, acquiring domain-agnostic states is essential for achieving efficient imitation learning. Without these states, imitation learning is hampered by domain-dependent information useless for control. However, existing methods fail to remove such disturbances from the states when the data from experts and agents show large domain shifts. To overcome this issue, we propose a domain-adversarial and -conditional state space model (DAC-SSM) that enables control systems to obtain domain-agnostic and task- and dynamics-aware states. DAC-SSM jointly optimizes the state inference, observation reconstruction, forward dynamics, and reward models. To remove domain-dependent information from the states, the model is trained with domain discriminators in an adversarial manner, and the reconstruction is conditioned on domain labels. We experimentally evaluated the model predictive control performance via imitation learning for continuous control of sparse reward tasks in simulators and compared it with the performance of the existing SRL method. The agents from DAC-SSM achieved performance comparable to experts and more than twice the baselines. We conclude domain-agnostic states are essential for imitation learning that has large domain shifts and can be obtained using DAC-SSM.
\end{abstract}
%%%%%%%%%%%%%%%%%%%%%%%%%%%%%%%%%%%%%%%%%%%%%%%%%%%%%%%%%%%%%%%%%%%%%%%%%%%%%%%%

\section{Introduction}
  In the context of imitation learning, it is natural to assume that the data from experts and agents have domain shifts~\cite{torabi:IfO}. For achieving efficient imitation learning in partially observable Markov decision processes (POMDPs), it is essential to acquire domain-agnostic state representation from observations. Without the domain-agnostic states, imitation learning is hampered by domain-dependent information, e.g. appearance of robots, which is useless for control of the robots. However, current state representation learning (SRL) methods~\cite{lesort:SRL} fail to remove such disturbances from the states. In imitation learning, a discriminator serves as an imitation reward function to distinguish the state-action pairs of the experts from those of the agents~\cite{ho:GAIL}. If the obtained states are NOT domain-agnostic, the discriminator is disturbed by the domain-dependent information, which is eye-catching but unrelated to tasks. As a result, the imitation rewards become unsuitable for the control, and imitation learning will be disrupted.

\begin{figure}[t]
  \centering
    \begin{minipage}{0.45\textwidth}
      \centering
      \includegraphics[width=\textwidth]{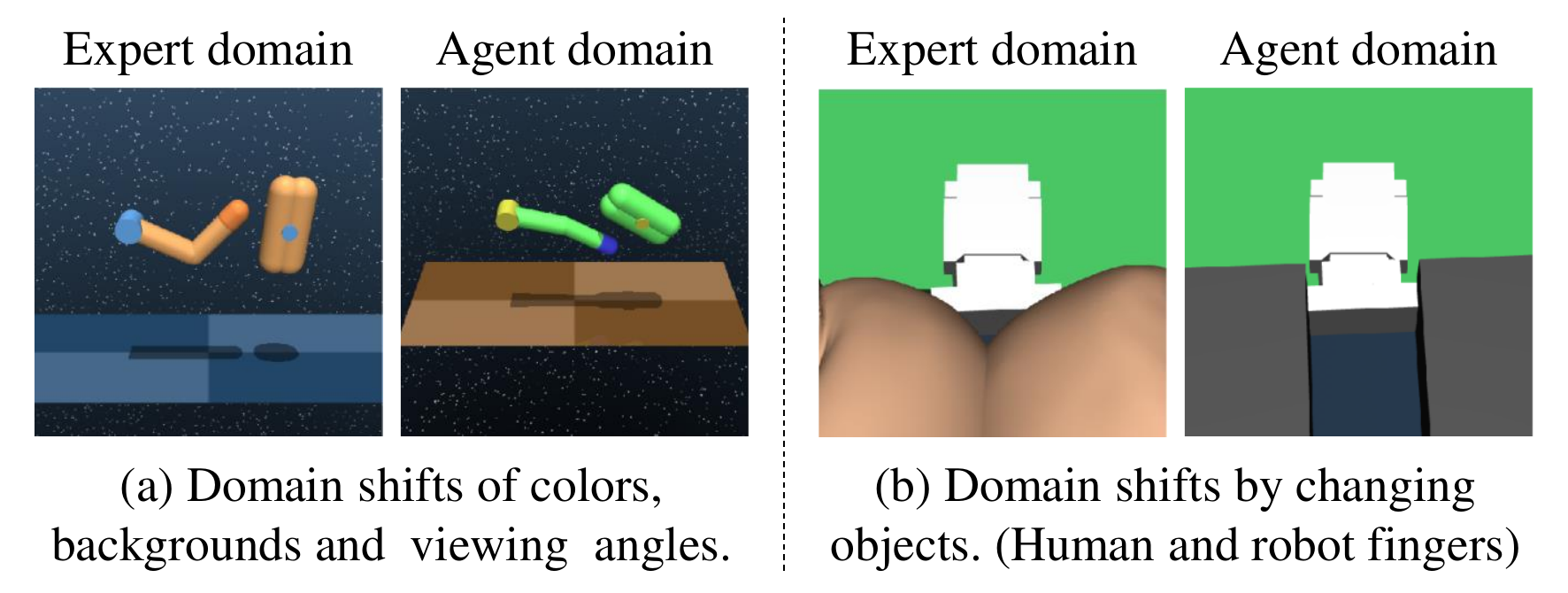}
    \end{minipage}
  \caption{Examples of domain shifts between experts and agents. We define the domain shifts as control-irrelevant changes in data like appearance. (a) Colors, backgrounds and viewing angles are different between the two domains. (b) The appearance of objects is different between the two domains. Human and robot fingers hold connectors in the expert and agent domain, respectively.}
\label{fig:domain-shifts}
\end{figure}

\begin{figure*}[t]
  \centering
  \begin{minipage}{0.32\textwidth}
    \centering
    \includegraphics[width=\textwidth]{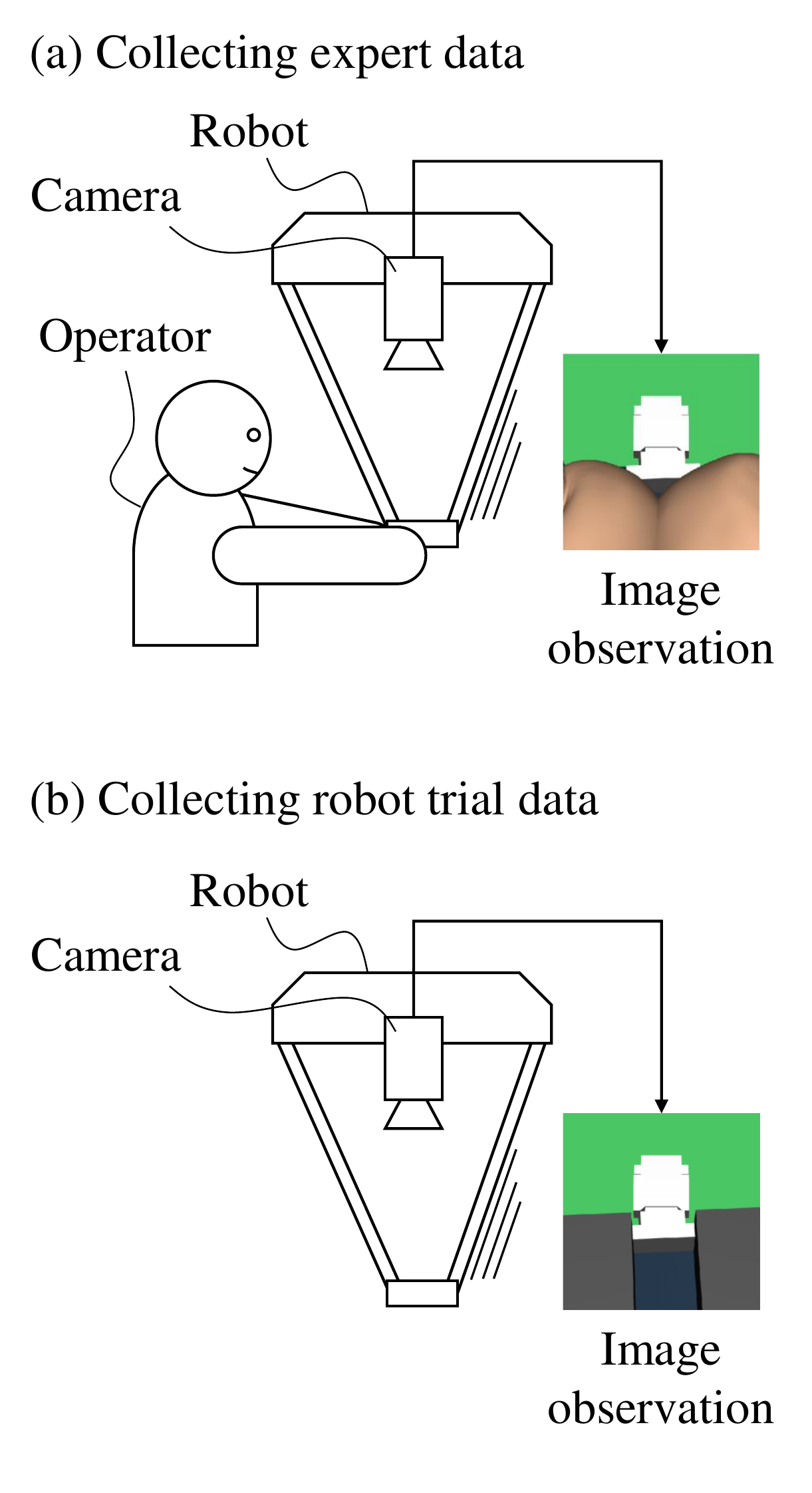}
  \end{minipage}
  \begin{minipage}{0.32\textwidth}
    \centering
    \includegraphics[width=\textwidth]{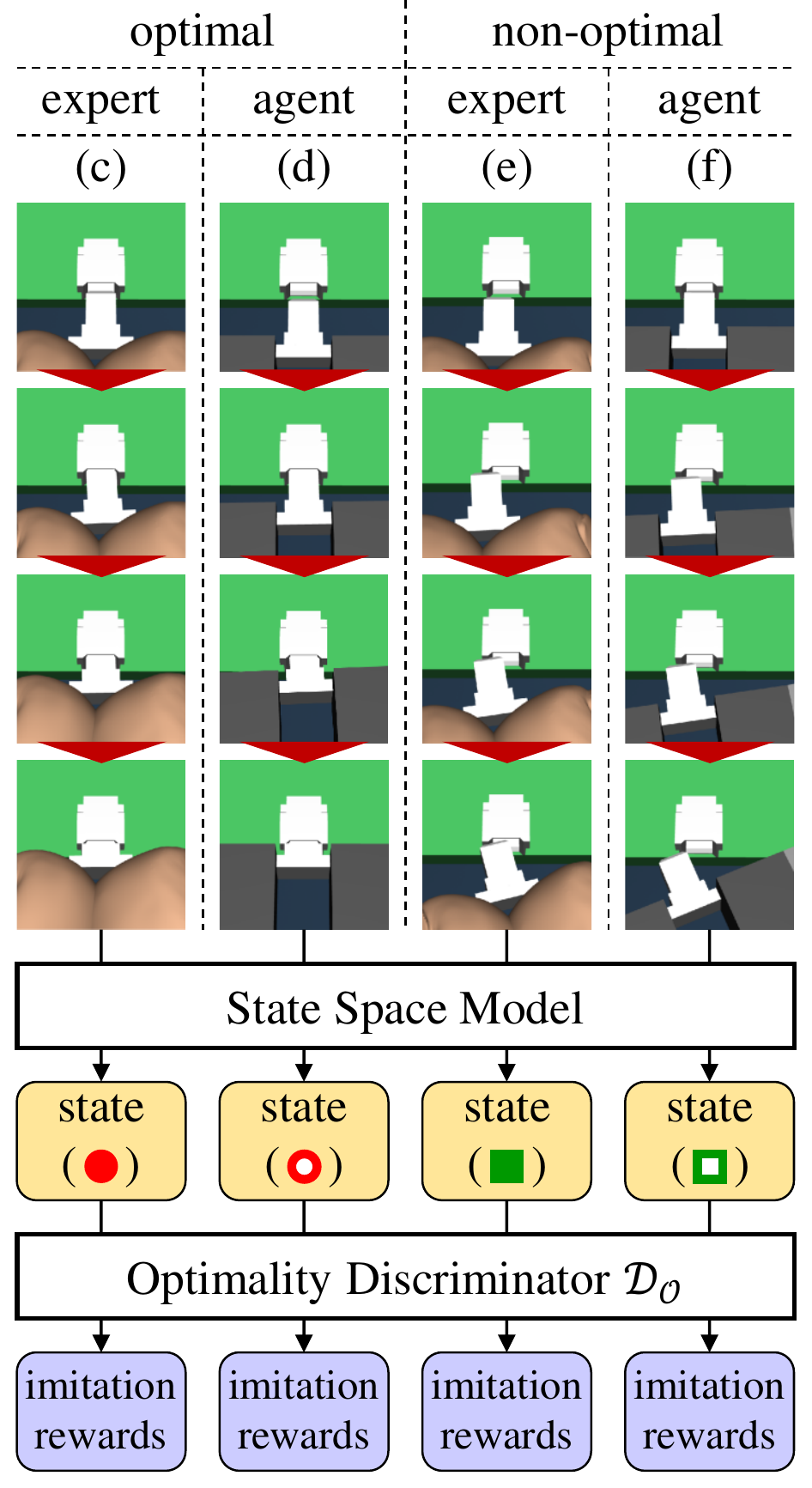}
  \end{minipage}
  \begin{minipage}{0.32\textwidth}
    \centering
    \includegraphics[width=\textwidth]{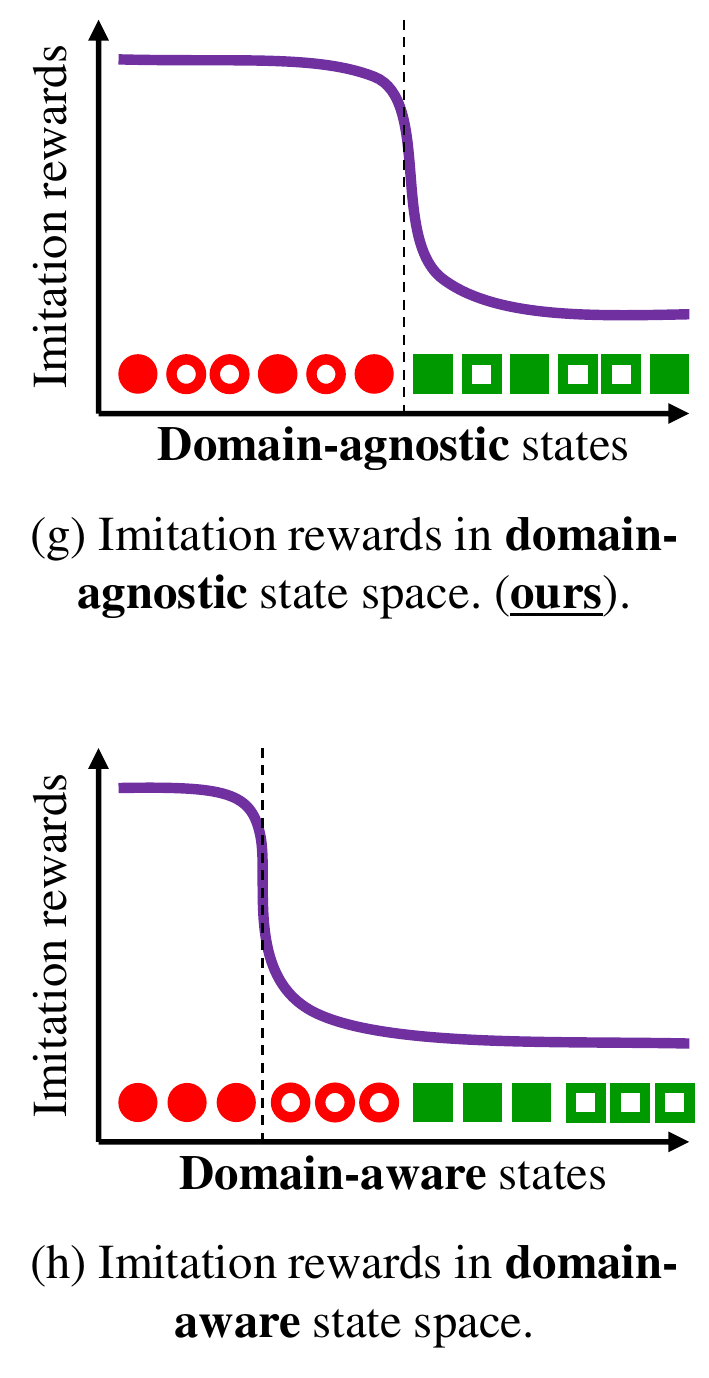}
  \end{minipage}
  \caption{Imitation rewards computed from image observations via domain-agnostic and -aware state space models. (a) and (b) show examples of expert image observations, and collection method of expert and agent data, respectively. In the expert images, human fingers hold a connector. In the agent images, robot fingers hold the connector. Therefore, appearance differs between the expert and agent domains. (c) and (d) show optimal trajectories of a connector insertion task from an expert and agent domain, respectively. (e) and (f) show non-optimal trajectories from the expert and agent domain, respectively. The middle diagram shows flow of the imitation rewards computation. Latent states are inferred from the image sequences via the state space models. Then the imitation rewards are calculated by an optimality discriminator $\mathcal{D}_\mathcal{O}$ from the obtained states. (g) and (h) show the imitation rewards from domain-agnostic and domain-aware states, respectively. In the domain-agnostic state space obtained via our DAC-SSM, the imitation rewards are high for the optimal trajectories, without relation to the domains. In the domain-aware state space, however, the imitation rewards are low for the optimal trajectories from the agent domain, because the expert and agent domains are distinguishable for the optimality discriminator.}
\label{fig:concept}
\end{figure*}

  Fig. \ref{fig:domain-shifts} shows examples of domain shifts between the data from an expert and agent. We define the domain shifts as control-irrelevant changes of the data like appearance: e.g. colors, textures,  backgrounds, viewing angles, and objects that are unrelated to the control. The domain shifts are, for example, caused by changing camera settings, location of data collection, appearance of the robot and so on. The domain shifts are also caused when unseen objects in one domain appear in the other domain. For example, an operator will be present in the expert images when she/he makes demonstrations via the direct teaching mode of a robot like Fig. \ref{fig:concept} (a) and (b). In this case, the existence of the operator in the images is the cause of the domain shifts.

  To overcome these domain shifts, in this paper, we propose a domain-agnostic and task- and dynamics-aware SRL model, called a domain-adversarial and -conditional state space model (DAC-SSM). Fig. \ref{fig:concept} shows how our DAC-SSM achieves the efficient imitation learning compared to the existing models. We pre-collected the expert and novice data, optimal and non-optimal trajectories in an expert domain, like Fig. \ref{fig:concept} (c) and (e). The agent data, like Fig. \ref{fig:concept} (d) and (f), are collected during training. Fig. \ref{fig:concept} (g) shows domain-agnostic state space obtained via DAC-SSM. Higher rewards are provided to the agents for optimal behavior, even when the agent data have large domain shifts from the expert data. Unlike DAC-SSM, the agents can not receive appropriate rewards in domain-aware state space via existing SRL method, like Fig. \ref{fig:concept} (h).
  
  DAC-SSM builds on a recurrent state space model (RSSM)~\cite{hafner2018planet}, and is trained with an optimality discriminator and a domain discriminator. The optimality discriminator serves as an imitation reward function. To remove the domain-dependent information from the states, (1) the state space is trained with the domain discriminator in an adversarial manner, and (2) the encoder and decoder of DAC-SSM are conditioned on domain labels. The domain discriminator is trained to identify which domain the acquired states belong to. The negative loss function of the domain discriminator, called the domain confusion loss~\cite{Tzeng2014DeepDC}, is added to the loss function of the state space. To reduce the domain confusion loss, the states are trained to be domain-agnostic. In other words, due to the domain confusion loss, DAC-SSM is trained to infer the states that have few clues for the domain discriminator to distinguish domain of the states. Moreover, the states are disentangled by conditional domain labels for the encoder and decoder, like conditional variational autoencoders (CVAE)~\cite{cvae}. Owing to the disentanglement, the domain-dependent information is eliminated from the state representation. Because DAC-SSM jointly optimizes the state inference, observation reconstruction, forward dynamics, and reward models, the obtained states are also task- and dynamics-aware as well as domain-agnostic.
  
  The main contribution of this work is a method to obtain the domain-agnostic states for imitation learning in POMDPs, using domain confusion loss and domain-conditional encoder and decoder. We experimentally compared our proposed model, DAC-SSM, to the existing SRL methods in terms of model predictive control (MPC) performance via imitation learning for continuous control sparse reward tasks of robots in the MuJoCo physics simulator~\cite{Todorov2012MuJoCoAP}. The agents in DAC-SSM achieved a performance comparable to the expert and more than twice that of the baselines. We also demonstrated that DAC-SSM successfully eliminated domain specific information like appearance from the states.

\section{Related studies}
  \subsection{State representation learning}
    SRL has been studied to obtain compact and expressive representation of robot control tasks from high-dimensional sensor data, such as images~\cite{lesort:SRL}. Appropriate state representation enables agents to achieve high performance for discrete and continuous control tasks from games~\cite{ha:worldmodels} to real robots~\cite{wang:infogan}. Sequential state space models, in which historical information of a control system is propagated via contextual states, have been shown to improve the performance and sample efficiency of robot control tasks. For example, previous work~\cite{hafner2018planet} proposed deep planning network (PlaNet), a planning methodology in the latent space obtained via RSSM. In RSSM, the obtained states are task- and dynamics-aware because the state inference, observation reconstruction, forward dynamics, and reward models are jointly optimized. Previous work~\cite{lee2019slac} also proposed a sequential state space model that is jointly optimized with a policy. Previous work~\cite{Gangwani2019belief} jointly optimized the optimality discriminator using policy, forward and inverse dynamics, and action models to obtain task- and dynamics-aware state representation. Their state representation, however, is not domain-agnostic.

  \subsection{Domain-agnostic feature representation}
    There are roughly two types of approaches to obtain domain-agnostic feature representation: domain-adversarial training and disentanglement. The domain-adversarial training is a simple and effective approach to extract feature representation which is unrelated to domains of data. Previous work~\cite{Tzeng2014DeepDC,adda} added the domain confusion loss to the loss function of the feature extractor. The domain adversarial training is also used for sim-to-real transfer learning~\cite{graspgan, Fang2018MultiTaskDA}. Similar approach is introducing a gradient reversal layer~\cite{Ganin2015DomainAdversarialTO} which back-propagates a negative gradient of the domain discriminator loss to the feature extractor.
    
    CVAE~\cite{cvae} is a simple and well-known disentanglement method. In CVAE, the encoder and decoder are conditioned on labels to disentangle label-related and -unrelated information in the latent spaces. Previous work~\cite{disentanglement_NIPS2018_7404} proposed cross-domain autoencoders to disentangle shared and exclusive features between two domains. Previous work~\cite{domainseparation} used private and shared encoders to disentangle domain-specific and -invariant components of the representation. They applied domain adversarial loss to train the shared encoder to extract the domain-invariant features. Thus, this method is combination of the domain-adversarial training and disentanglement. They demonstrated that the obtained features were useful for several downstream tasks like classification.

  \subsection{Imitation learning}
    Imitation learning~\cite{schaal:IL} is a powerful and accepted approach that makes the agents mimic expert behavior by using a set of demonstrations of tasks. Previous work~\cite{ho:GAIL} proposed an imitation learning framework called Generative Adversarial Imitation Learning (GAIL). In GAIL, imitation rewards are computed by the optimality discriminator, which distinguishes if a state-action pair is generated by an agent policy or from the expert demonstrations. They formulated a joint process of reinforcement learning and inverse reinforcement learning as a two-player game of the policy and discriminator, analogous to Generative Adversarial Nets~\cite{goodfellow:GAN}. GAIL has been shown to solve complex high-dimensional continuous control tasks~\cite{Kostrikov2018DiscriminatorActorCriticAS, pmlr-v70-baram17a, Li2017InfoGAILII, Sharma2018DirectedInfoGL}.
    
  \subsection{Imitation learning with the domain shifts}
    Using common measurable features is one of the popular approaches for the imitation learning with the domain shifts. For example, keypoints of objects~\cite{sieb:graph} and marker positions~\cite{gupta2016learning,lee:silo} are tracked as the states. In these approaches, one can directly apply existing imitation learning techniques without focusing on the domain shifts. However, such features are not always available.
    Previous work~\cite{Stadie2017ThirdPersonIL} added the domain confusion loss to the optimality discriminator to make it domain-agnostic. By computing the imitation reward using the discriminator, they successfully achieved imitation learning from third-person perspective images. Their approach, however, does not include SRL, and suffered from sample inefficient nature of model-free reinforcement learning.

  \subsection{Imitation Learning from Observation}
    Imitation learning from observation (IfO) assumes that demonstration data have only states or observations~\cite{torabi:IfO}. In IfO, action information is not observable in the demonstration data. It is more practical problem setting than imitation learning, but harder to achieve precise tasks. Previous work~\cite{Merel2017LearningHB} achieved IfO from human motion capture data. They extracted hand-designed features from human and robot data to absorb difference between their body structures. Their method, however, heavily depends on the feature design. Previous work~\cite{LiuIfO2018} proposed a context translation model to predict a desired future observation from a pair of target observations at the first timestep and demonstrated source observations. The agent was trained to generate observations that match the predicted desired observations. The context translation model, however, requires pairs of source and target trajectories carefully aligned along timesteps. Previous work~\cite{tcn} trained a feature extractor from images using contrastive learning. They selected positive images from the same timestep in different viewpoints, and negative ones from distant timesteps in the same viewpoint. Thus, they obtained viewpoint-agnostic states. These works enabled their agents to roughly imitate human demonstration, but they did not apply their method to more complex and highly accurate control tasks. Previous work~\cite{Smith2019AVIDLM} used Cycle-GAN~\cite{cyclegan} to generate instruction images in the agent domain from human demonstration. The generated images were used to calculate rewards for MPC. Their approach, however, requires a large amount of human and robot data beforehand.

    %Another approach is a training domain-adaptive policy \cite{yu:maml,yu:HIL} via meta-learning \cite{finn2017model}.
    
\section{Proposed Method} \label{sec:method}
  In this section, we explain DAC-SSM and planning algorithm. DAC-SSM is a state space model to obtain domain-agnostic states by introducing combination of simple domain adversarial training and disentanglement. In Sec.~\ref{sec:dac-ssm}, we formulate our models of our proposed method. In Sec.~\ref{sec:discriminator}, we describe the domain and optimality discriminators. In Sec.~\ref{sec:training}, we further explain the training architecture of DAC-SSM. In Sec.~\ref{sec:mpc}, we describe the planning algorithm.

\begin{figure}[t]
  \centering
    \begin{minipage}{0.3\textwidth}
      \centering
      \includegraphics[width=\textwidth]{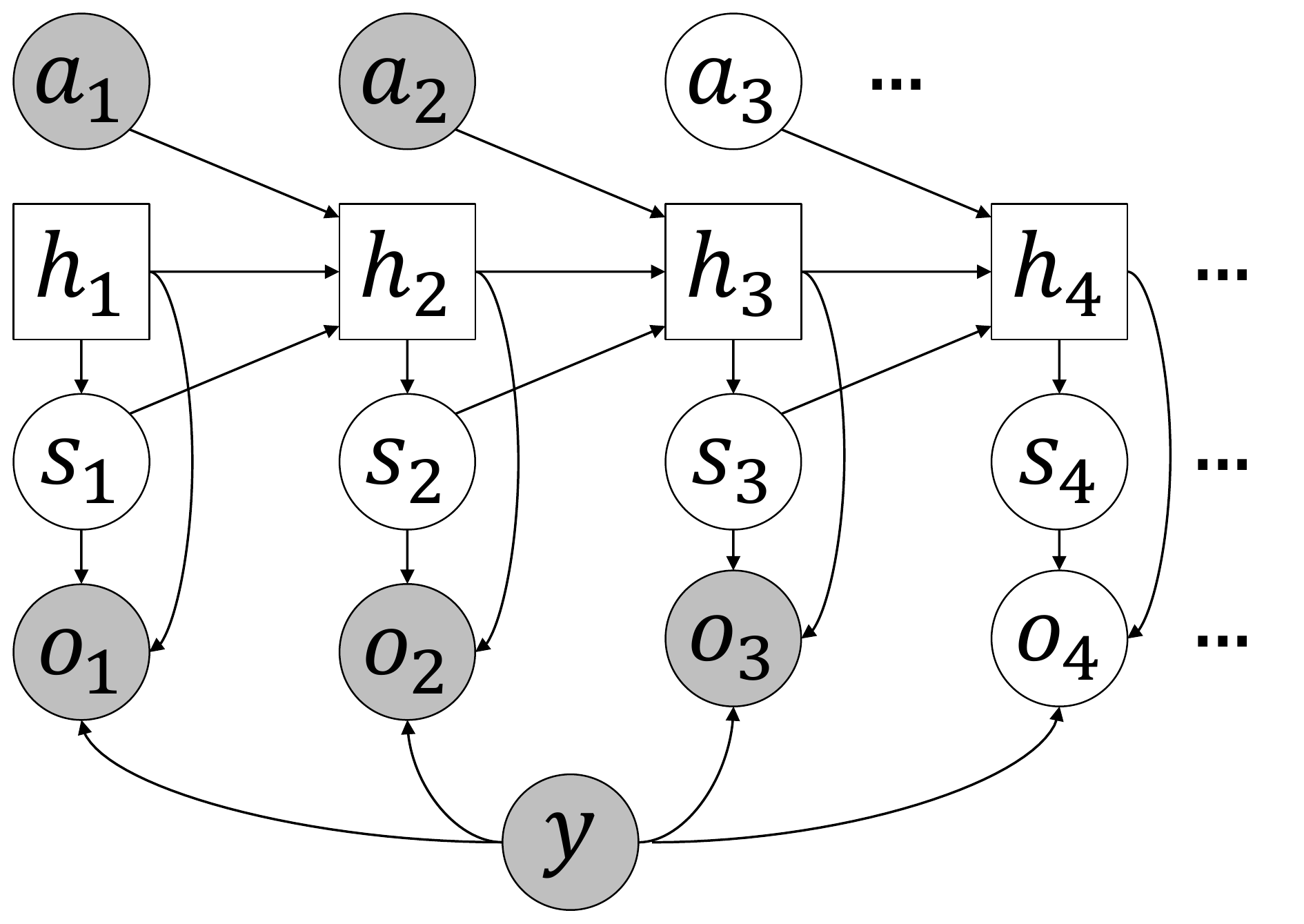}
    \end{minipage}
  \caption{Graphical model of our DAC-SSM. A discrete time step: $t$, contextual deterministic states: $h_t$, stochastic states: $s_t$, image observations: $o_t$, continuous actions: $a_t$, and domain labels: $y$.}
\label{fig:graphical_model}
\end{figure}

  \subsection{State space model} \label{sec:dac-ssm}
    Fig.~\ref{fig:graphical_model} shows a graphical model of our DAC-SSM. It is modeled as POMDPs. In POMDPs, an individual image does not have all the information about the states. Therefore, our model builds on RSSM, which has contextual states to propagate historical information. We use the following notations: a discrete time step, $t$, contextual deterministic states, $h_t$, stochastic states, $s_t$, image observations, $o_t$, continuous actions, $a_t$, and domain labels, $y$. The model follows the mixed deterministic/stochastic dynamics below:
    \begin{itemize}
      \setlength{\parskip}{0.05cm}
      \setlength{\itemsep}{0.05cm}
      \item Transition model: $h_t = f(h_{t-1}, s_{t-1}, a_{t-1})$
      \item State model: $s_t \sim p(s_t | h_t)$
      \item Observation model: $o_t \sim p(o_t | h_t, s_t, y)$
    \end{itemize}
    Transition model $f(h_{t-1}, s_{t-1}, a_{t-1})$ was implemented as a recurrent neural network. To train the model, we maximized the probability of a sequence of observations in the entire generative process:
    \begin{equation}
      \begin{split}
        p(o_{1:T}|a_{1:T},y) =& \int \prod_t \Bigl[ p(s_t|h_t) p(o_t|h_t,s_t,y) \Bigr] ds_{1:T} \\
        where \quad h_t=&f(h_{t-1},s_{t-1},a_{t-1})
      \end{split}
    \end{equation}
    Generally this objective is intractable. We utilize the following evidence lower bound (ELBO) on the log-likelihood by introducing the posterior $q(s_t|o_{\leq t},a_{<t},y)$ to infer the approximate stochastic states.
    \begin{equation}
      \begin{split}
        \ln p&(o_{1:T}|a_{1:T},y) \\
        \geq & \sum_{t=1}^T \mathbb{E}_{q(s_t|o_{\leq t},a_{<t},y)} \bigl[ \ln p(o_t|h_t,s_t,y) \bigr] \\
        & \;\; - \mathbb{E}_{q(s_{t-1}|o_{\leq t-1},a_{<t-1},y)} \bigl[ {\rm KL} [q(s_t|o_{\leq t},a_{<t},y)||p(s_t|h_t)] \bigr] \\
        =& -\mathcal{L}_{\rm RSSM}
      \end{split}
    \end{equation}
    The posterior $q(s_t|o_{\leq t},a_{<t},y)$ and the observation model $p(o_t | h_t, s_t, y)$ are implemented as an encoder and decoder, respectively. They are conditioned on the domain labels, $y$. The domain labels help them to change their behavior depending on the domain. The domain-dependent information is eliminated from the obtained states $s_t$ and $h_t$, like CVAE.

  \subsection{Domain and optimality discriminators} \label{sec:discriminator}
    We further introduce the domain and optimality discriminators, $\mathcal{D}_d$ and $\mathcal{D}_\mathcal{O}$. The role of the domain discriminator is computing the domain confusion losses. The optimality discriminator serves as an imitation reward function. We utilized three types of datasets: the expert, novice and agent data. The expert data are successful trajectories in the expert domain, whereas the novice data are non-optimal trajectories in the expert domain. We pre-collected the expert and novice data. By using both of them, the domain discriminator becomes agnostic on the optimality because these data are in the same domain but have different optimality. Similarly, the optimality discriminator becomes agnostic on the domain, but aware of the optimality. The agent data are collected during training. We denote replay buffers for the data from the agents, experts, and novices as $\mathcal{B}_A$, $\mathcal{B}_E$, and $\mathcal{B}_N$, respectively. The loss function of the domain discriminator is denoted as follows:
    \begin{equation}
      \begin{split}
        \mathcal{L_D}_d =& 2 \times \mathbb{E}_{h_t \sim \mathcal{B}_A} [\ln \mathcal{D}_d (h_t)] \\
        &+ \mathbb{E}_{h_t \sim \mathcal{B}_E} [\ln (1-\mathcal{D}_d(h_t))] \\
        &+ \mathbb{E}_{h_t \sim \mathcal{B}_N} [\ln (1-\mathcal{D}_d(h_t))]
      \end{split}
    \end{equation}
    Here, we introduce a simple abbreviation of the expectation to avoid complexity: 
    \begin{equation}
      \begin{split}
        \mathbb{E}_{h_t \sim \mathcal{B}} [\cdot] \equiv & \mathbb{E}_{o_{\leq t-1}, a_{\leq t-1}, y\sim \mathcal{B}} \;
        \mathbb{E}_{\substack{s_{t-1} \sim q(s_{t-1}|o_{\leq t-1},a_{<t-1},y) \\ h_t=f(h_{t-1},s_{t-1},a_{t-1})}} [\cdot] \\
      \end{split}
    \end{equation}
    Similarly, the loss function of the optimality discriminator is denoted as follows:
    \begin{equation}
      \begin{split}
        \mathcal{L_{D_{\mathcal{O}}}} =& \mathbb{E}_{h_t, a_t \sim \mathcal{B}_A} [\ln \mathcal{D}_{\mathcal{O}} (h_t,a_t)] \\
        &+ 2 \times \mathbb{E}_{h_t, a_t \sim \mathcal{B}_E} [\ln (1-\mathcal{D}_{\mathcal{O}}(h_t,a_t))] \\
        &+ \mathbb{E}_{h_t, a_t \sim \mathcal{B}_N} [\ln \mathcal{D}_{\mathcal{O}}(h_t,a_t)]
      \end{split}
    \end{equation}
    It is trained to distinguish if state-action pairs $(h_t,a_t)$ are from episodes of the experts or not.

  \subsection{Training of DAC-SSM} \label{sec:training}
\begin{figure}[t]
  \centering
  \begin{minipage}{0.45\textwidth}
    \centering
    \includegraphics[width=\textwidth]{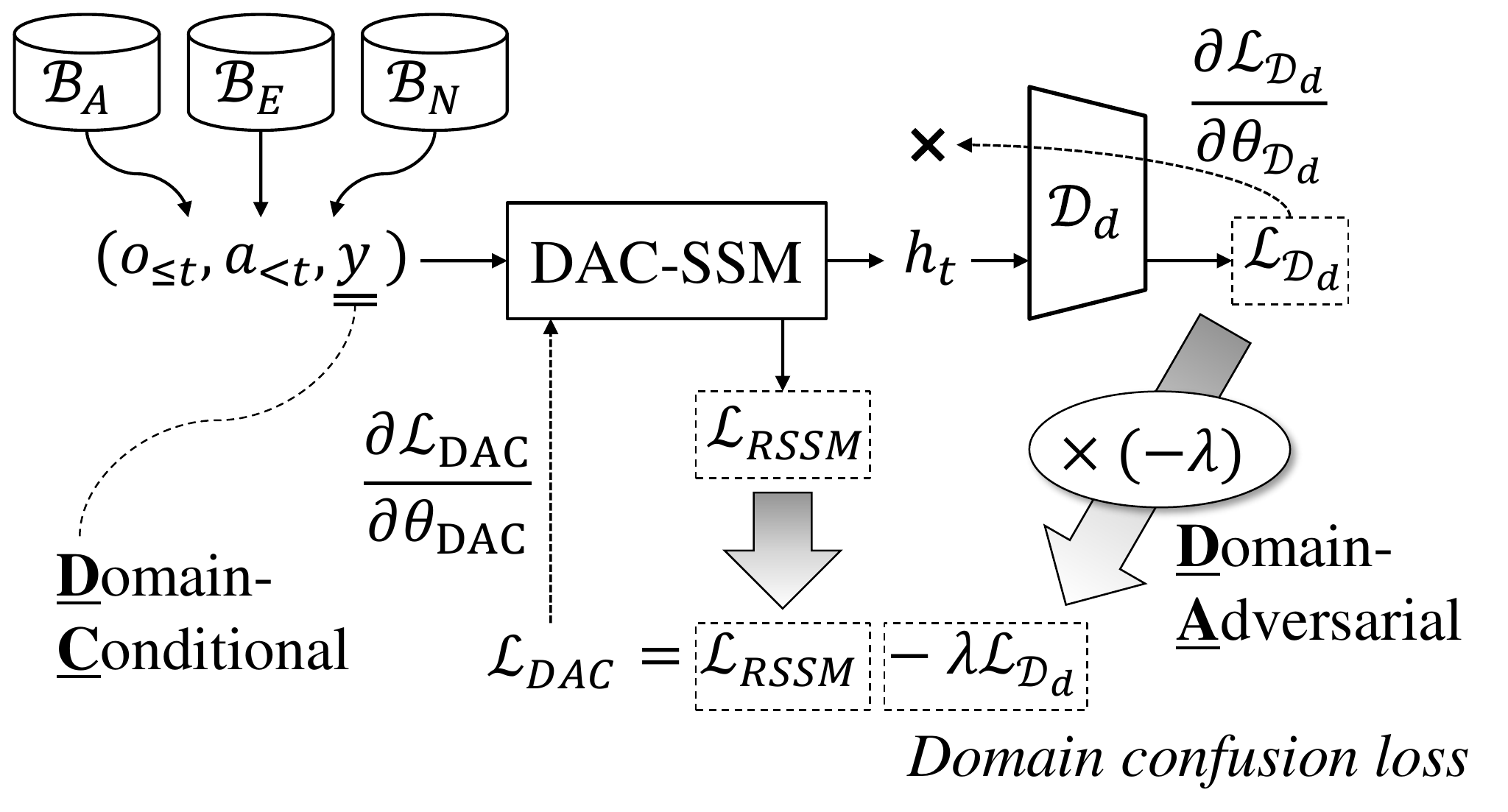}
  \end{minipage}
  \caption{Training architecture of DAC-SSM. The dashed lines represent back-propagation paths. The domain confusion losses $-\lambda \mathcal{L_D}_d$ are added to the state space losses $\mathcal{L}_{RSSM}$. $\mathcal{B}_A$, $\mathcal{B}_E$, and $\mathcal{B}_N$ represent replay buffers for the data from the agents, experts, and novices. $\mathcal{D}_d$ represents the domain discriminator.}
  \label{fig:training-architecture}
\end{figure}

    Fig. \ref{fig:training-architecture} displays a diagram of training architecture of DAC-SSM. The dashed lines represent back-propagation paths. The model is trained by minimizing state space losses with the domain confusion losses:
    \begin{equation} \label{eq:DAC-SSM}
      \mathcal{L}_{DAC}=\mathcal{L}_{RSSM}-\lambda \mathcal{L_D}_d
    \end{equation}
    where $\lambda$ is a hyper-parameter. The reward models, $r_t \sim p(r_t|h_t,s_t)$, are trained by the losses:
    \begin{equation} \label{eq:reward_model}
      \mathcal{L}_r = -\sum_{t=1}^T \mathbb{E}_{q(s_t|o_{\leq t},a_{<t},y)} \bigl[ \ln p(r_t|h_t,s_t) \bigr]
    \end{equation}
    The gradient of the optimality discriminator losses, ${\partial \mathcal{L_{D_{\mathcal{O}}}}}/{\partial \theta_{\mathcal{D}_{\mathcal{O}}}}$, is not propagated to DAC-SSM. On the other hand, the gradient of the domain discriminator losses, ${\partial \mathcal{L_D}}_d/{\partial \theta_\mathcal{D}}_d$, is not directly propagated to DAC-SSM, but the domain confusion losses, $-\lambda \mathcal{L_D}_d$, are added to the state space losses, $\mathcal{L}_{RSSM}$.
    Thus, the obtained states become domain-agnostic, and task- and dynamics-aware. Therefore, the states have considerable information that is useful for control (task- and dynamics-aware), but few clues regarding the domain-dependent information (domain-agnostic).

  \subsection{Planning algorithm} \label{sec:mpc}
    We used MPC~\cite{Garcia:1989:MPC:72068.72069, Okada2019VariationalIM, Okada2017PathIN} for planning in the obtained state space via DAC-SSM. For planning algorithm of MPC, cross entropy method (CEM)~\cite{boer:cem} was used to search for the best action sequences. CEM is a robust population-based approach to infer an optimal distribution over action sequences. The action sequences are optimized to maximize an objective. Modeling the objective as a function of the states and actions makes computational cost of the planning much lighter. The planning is executed purely in the low-dimensional latent space without generating images~\cite{hafner2018planet, dreamer, planetofbayesians}. Multiple types of rewards are used for the objective~\cite{kinose2019, kaushik:hal-01884294} in the context of control as inference~\cite{levine2018reinforcement}. We define the distribution over the task-optimality, $\mathcal{O}^R_t$, as follows:
    \begin{equation}
      p(\mathcal{O}^R_t=1 | h_t, s_t) = \exp{(\mathbb{E}_{p(r_t|h_t, s_t)}[r_t])}
    \end{equation}
    The distribution over the imitation-optimality, $\mathcal{O}^I_t$, is calculated by using the optimality discriminator:
    \begin{equation}
      p(\mathcal{O}^I_t=1 | h_t, a_t) = \exp(\ln \mathcal{D}_{\mathcal{O}}(h_t, a_t)) = \mathcal{D}_{\mathcal{O}}(h_t,a_t)
    \end{equation}
    We use $h_t$ to calculate both rewards because contextual information is essential for the POMDPs. Hence, the objective of the CEM is to maximize the probability of the task- and imitation-optimalities, as given  below:
    \begin{equation}
      \begin{split}
        & \ln p(\mathcal{O}^R_{1:H}=1,\mathcal{O}^I_{1:H}=1|h_t,s_t,a_t) \\
        =& \sum^H_{t=1} \bigl[\mathbb{E}_{p(r_t|h_t, s_t)}[r_t] + \ln \mathcal{D}_{\mathcal{O}}(h_t,a_t) \bigr]
      \end{split}
    \end{equation}
    where $H$ is the planning horizon of CEM.

\section{Experiments} \label{sec:experiments}

\begin{figure*}[t]
  \centering
  \begin{minipage}{1.0\textwidth}
  \centering
    \includegraphics[width=\textwidth]{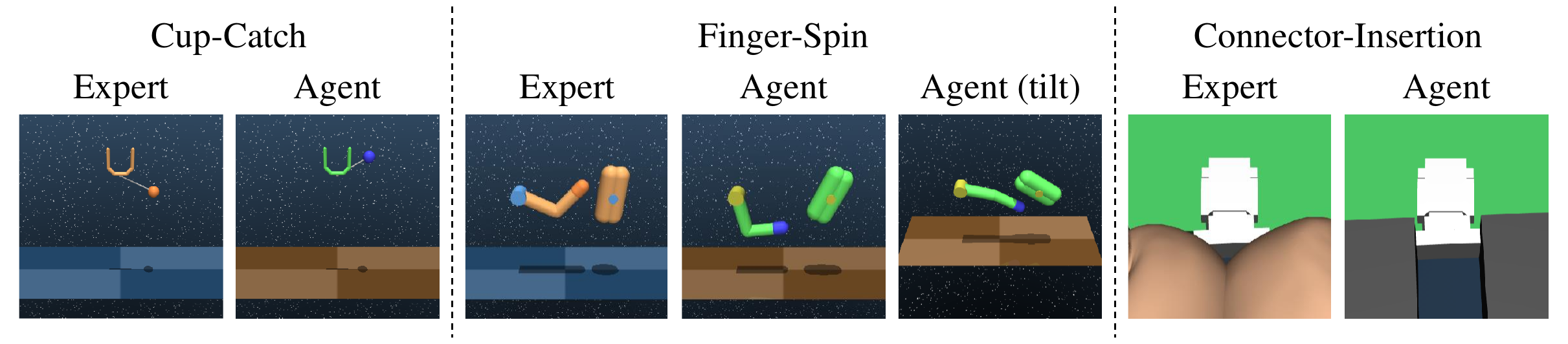}
  \end{minipage}
  \caption{We consider three tasks: Cup-Catch, Finger-Spin, and Connector-Insertion. Each task has expert and agent domain versions. We make two different agent domains for Finger-Spin. In one agent domain, color of objects and floors is different from the expert domain. In the other agent domain, viewing angles are further different. In the Connector-Insertion, human fingers hold the connector in the expert domain, while robot fingers hold it in the agent domain.}
  \label{fig:envs}
\end{figure*}

\begin{figure*}[t]
  \centering
  \begin{minipage}{1.0\textwidth}
    \begin{minipage}{0.24\textwidth}
      \centering Cup-Catch
    \end{minipage}
    \begin{minipage}{0.24\textwidth}
      \centering Finger-Spin
    \end{minipage}
    \begin{minipage}{0.24\textwidth}
      \centering Finger-Spin (Tilted view)
    \end{minipage}
    \begin{minipage}{0.24\textwidth}
      \centering Connector-Insertion
    \end{minipage}
  \end{minipage}
  \begin{minipage}{1.0\textwidth}
    \begin{minipage}{0.24\textwidth}
      \centering
      \includegraphics[width=\textwidth]{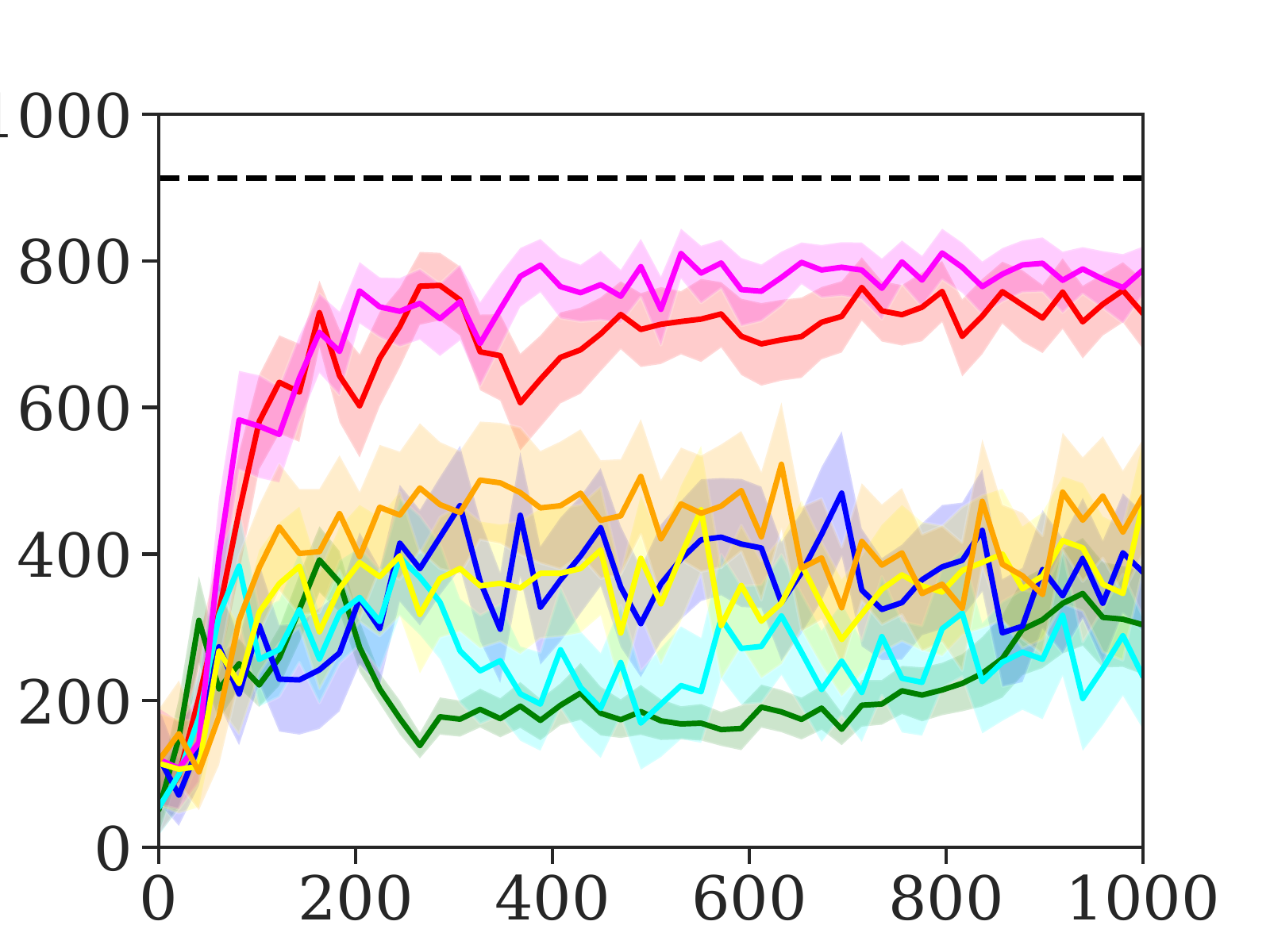}
    \end{minipage}
    \begin{minipage}{0.24\textwidth}
      \centering
      \includegraphics[width=\textwidth]{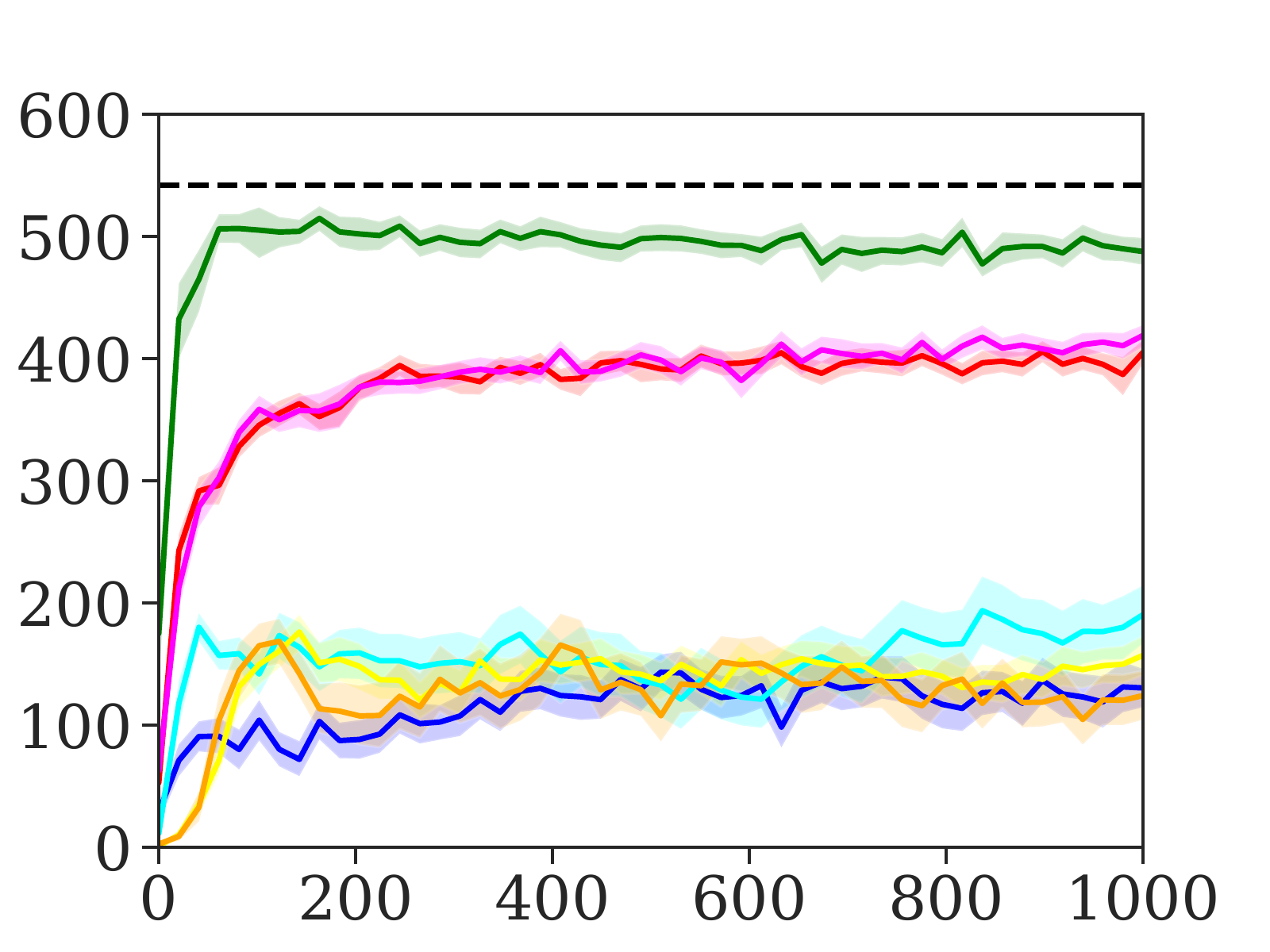}
    \end{minipage}
    \begin{minipage}{0.24\textwidth}
      \centering
      \includegraphics[width=\textwidth]{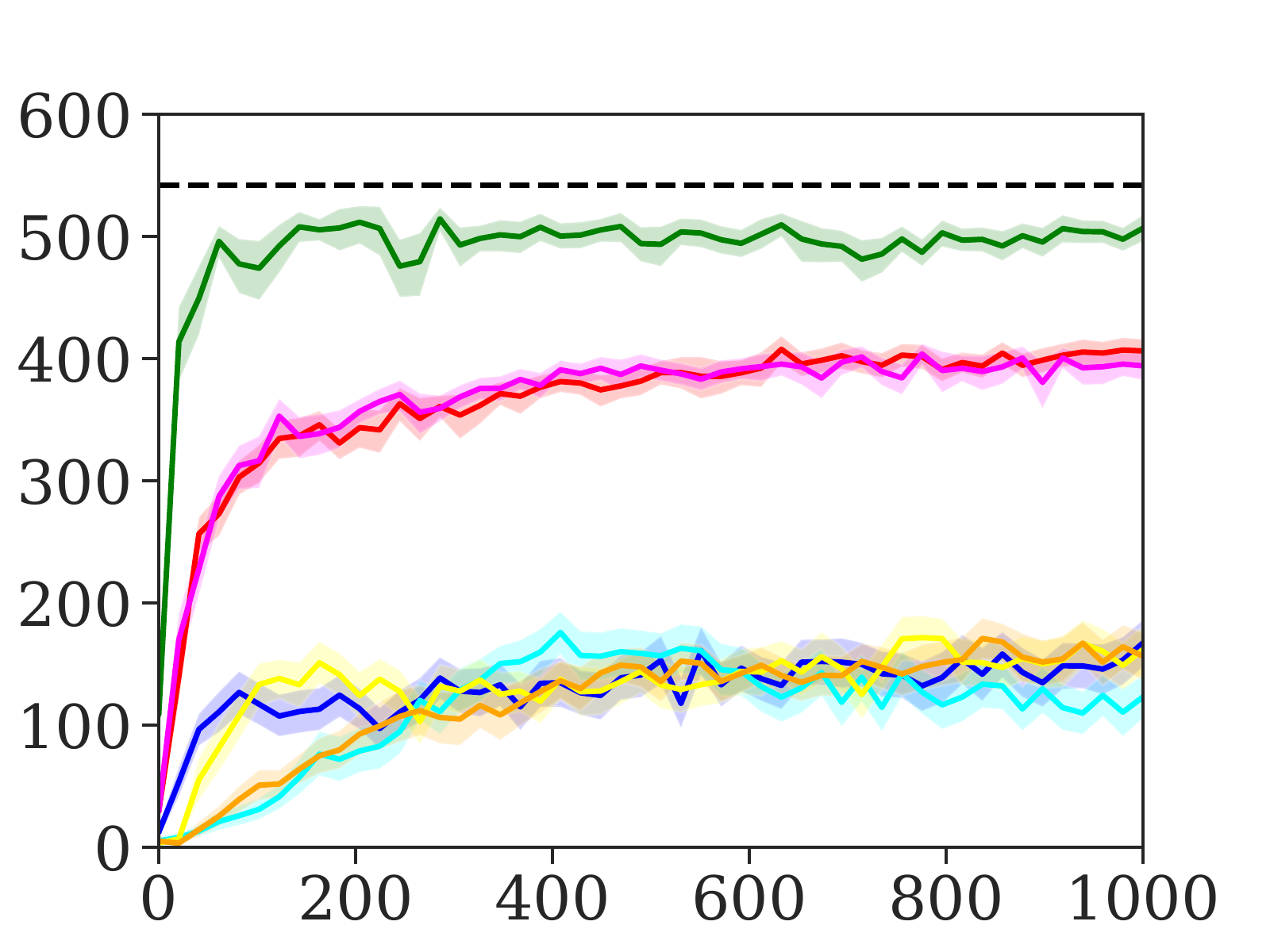}
    \end{minipage}
    \begin{minipage}{0.24\textwidth}
      \centering
      \includegraphics[width=\textwidth]{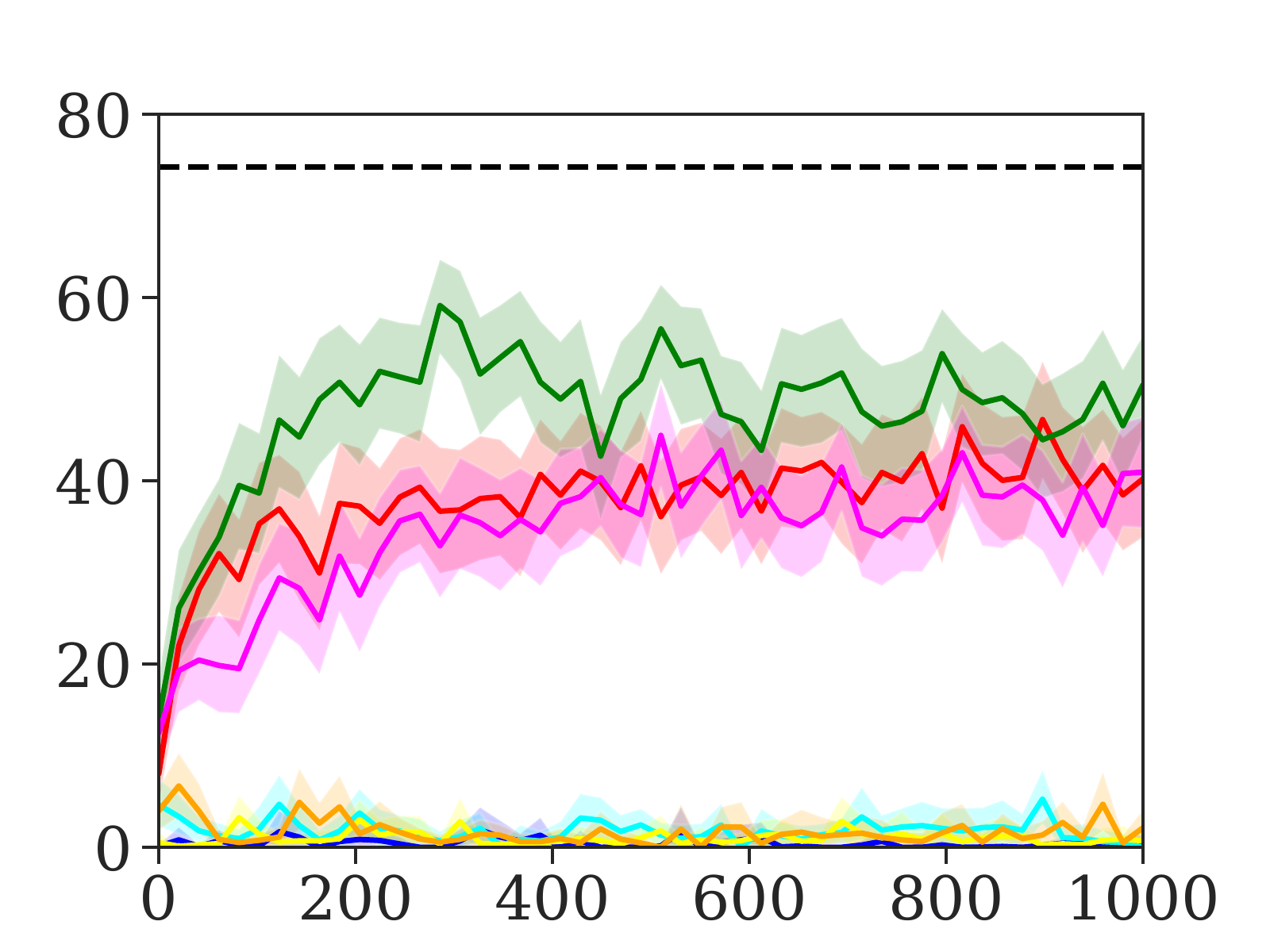}
    \end{minipage}
  \end{minipage}
  \begin{minipage}{1.0\textwidth}
    \centering
    \includegraphics[width=\textwidth]{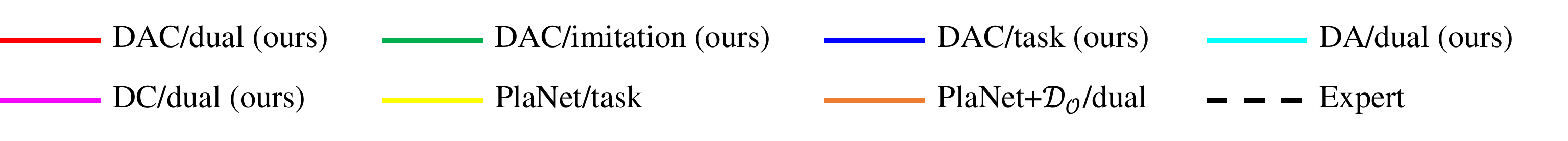}
  \end{minipage}
  \caption{Comparison of our proposed method with the baselines. The plots show the test performance over the number of collected trajectories by the agent. The solid lines show the means, and the colored areas show the percentiles from 5 to 95 over 20 trajectories across 4 seeds. The dashed lines show the average scores of the expert trajectories. We compare DAC-SSM with three types of reward function: \textit{task}, \textit{imitation} and \textit{dual}. The \textit{dual} means weighted sum of the \textit{task} and \textit{imitation} rewards. DAC-SSM: with the domain confusion loss and DC decoder. DA-SSM: with the domain confusion loss without the DC decoder. DC-SSM: without the domain confusion loss with the DC decoder. We used not only a DC decoder but also a DC encoder for the Finger-Spin of the tilted view. PlaNet+$\mathcal{D}_\mathcal{O}$: naive implementation of the optimality discriminator with RSSM.}
  \label{fig:dac-score}
\end{figure*}

\begin{table*}[t]
  \centering
  \caption{Mean MPC performance after 1,000 episodes. $\pm$ represents one standard deviation. Boldface indicates the best results. Underlines mean the second-best.}
  \begin{tabular}{c||ccccc|cc}
    \hline
    Task & DAC/dual & DAC/imitation & DAC/task & DA/dual & DC/dual & PlaNet & PlaNet+$\mathcal{D}_\mathcal{O}$ \\
    \hline
      Cup-Catch
      &\underline{728$\pm$223}&304$\pm$323&375$\pm$371&233$\pm$350&\textbf{788$\pm$149}&470$\pm$398&479$\pm$359 \\
      Finger-Spin
      &405$\pm$42&\textbf{488$\pm$50}&130$\pm$73&190$\pm$108&\underline{419$\pm$41}&157$\pm$73&124$\pm$91 \\
    \begin{tabular}{c}
      Finger-Spin (tilted view)
    \end{tabular}
      &\underline{406$\pm$45}&\textbf{507$\pm$48}&167$\pm$87&123$\pm$80&394$\pm$51&162$\pm$87&156$\pm$89 \\
    \begin{tabular}{c}
      Connector-Insertion
    \end{tabular}
      &40.2$\pm$29.1&\textbf{50.5$\pm$25.0}&0.4$\pm$4.0&0.0$\pm$0.0&\underline{40.9$\pm$26.7}&0.7$\pm$3.4&2.1$\pm$8.1 \\
    \hline
  \end{tabular}
  \label{tab:score}
\end{table*}

\begin{figure*}[t]
  \centering
    \begin{minipage}{0.8\textwidth}
      \centering
      \includegraphics[width=\textwidth]{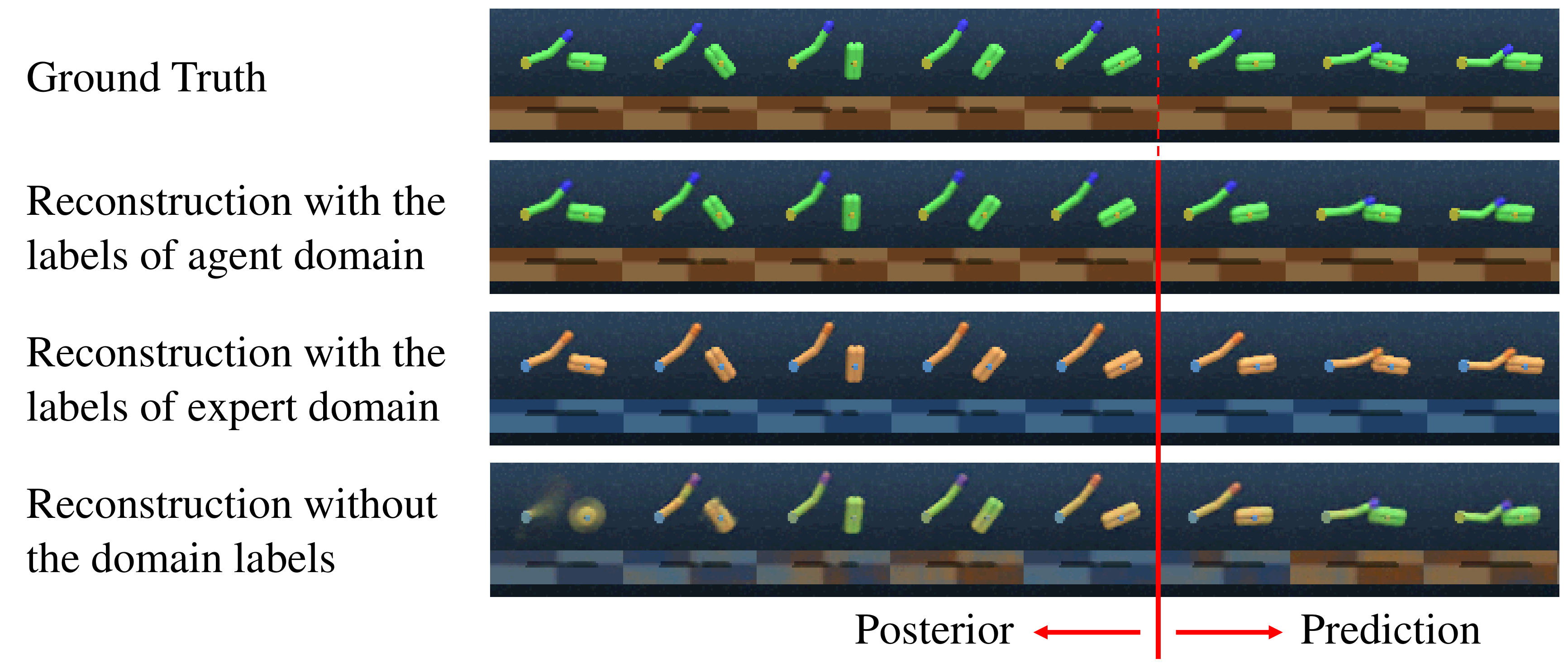}
  \end{minipage}
  \caption{Example image sequence (the first row) and corresponding open-loop video predictions (second to the last row) observed for the Finger-Spin task. Columns 1-5 are context frames and were reconstructed from posterior samples, and the remaining images were generated from open-loop prior samples. The second and third row was reconstructed with expert and agent domain labels, respectively. The last row was reconstructed from the contextual states, $h_t$, \textit{without} domain labels. Another decoder was trained separately for the reconstruction of the images in the last row. The first column of the last row was reconstructed from $h_t$ initialized by zero.}
  \label{fig:reconstruction}
\end{figure*}

\begin{figure}[t]
  \centering
  \begin{minipage}{0.35\textwidth}
      \centering
      \includegraphics[width=\textwidth]{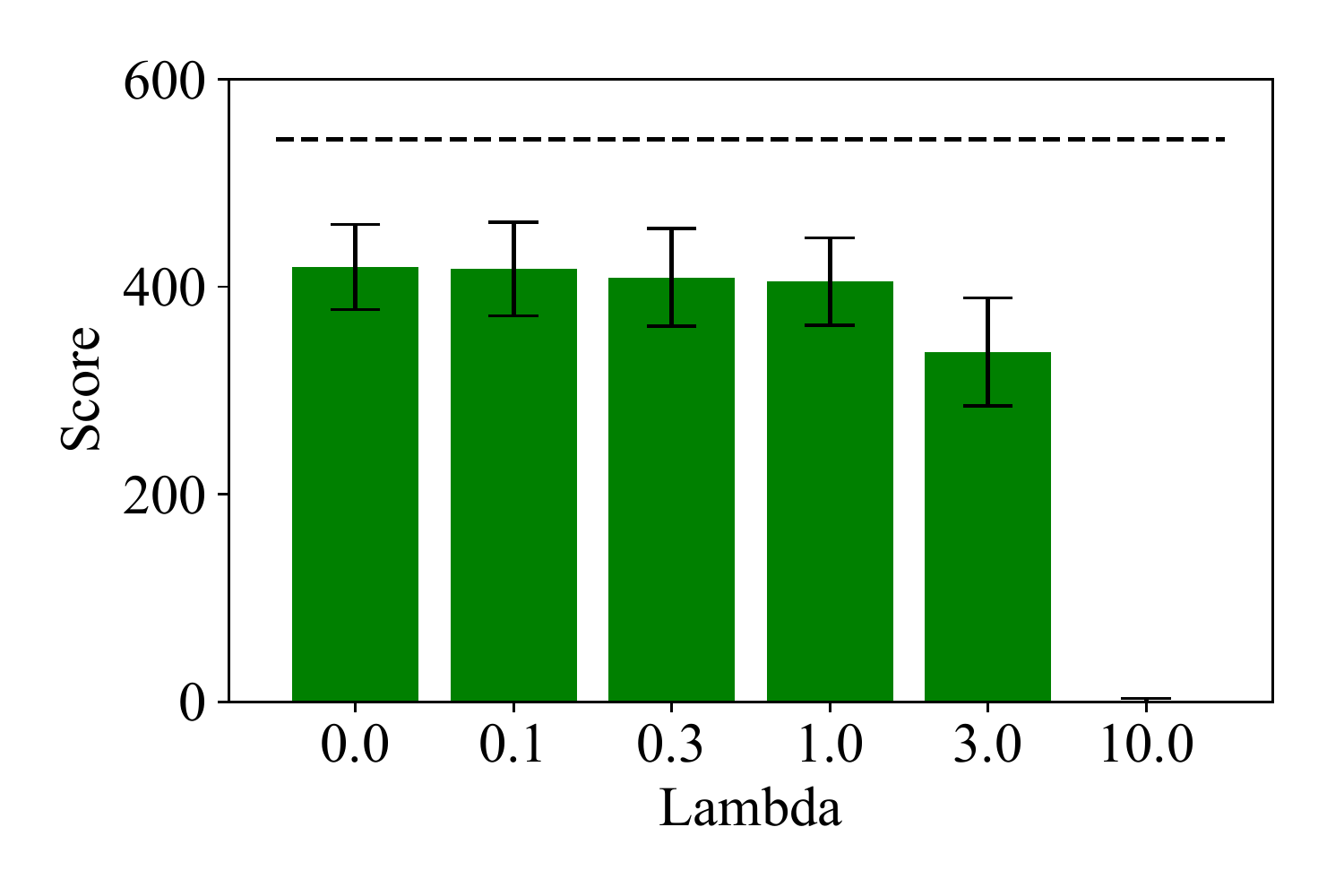}
    \end{minipage}
  \begin{minipage}{0.45\textwidth}
    \centering (a) Finger-Spin
  \end{minipage}
  \begin{minipage}{0.35\textwidth}
    \centering
    \includegraphics[width=\textwidth]{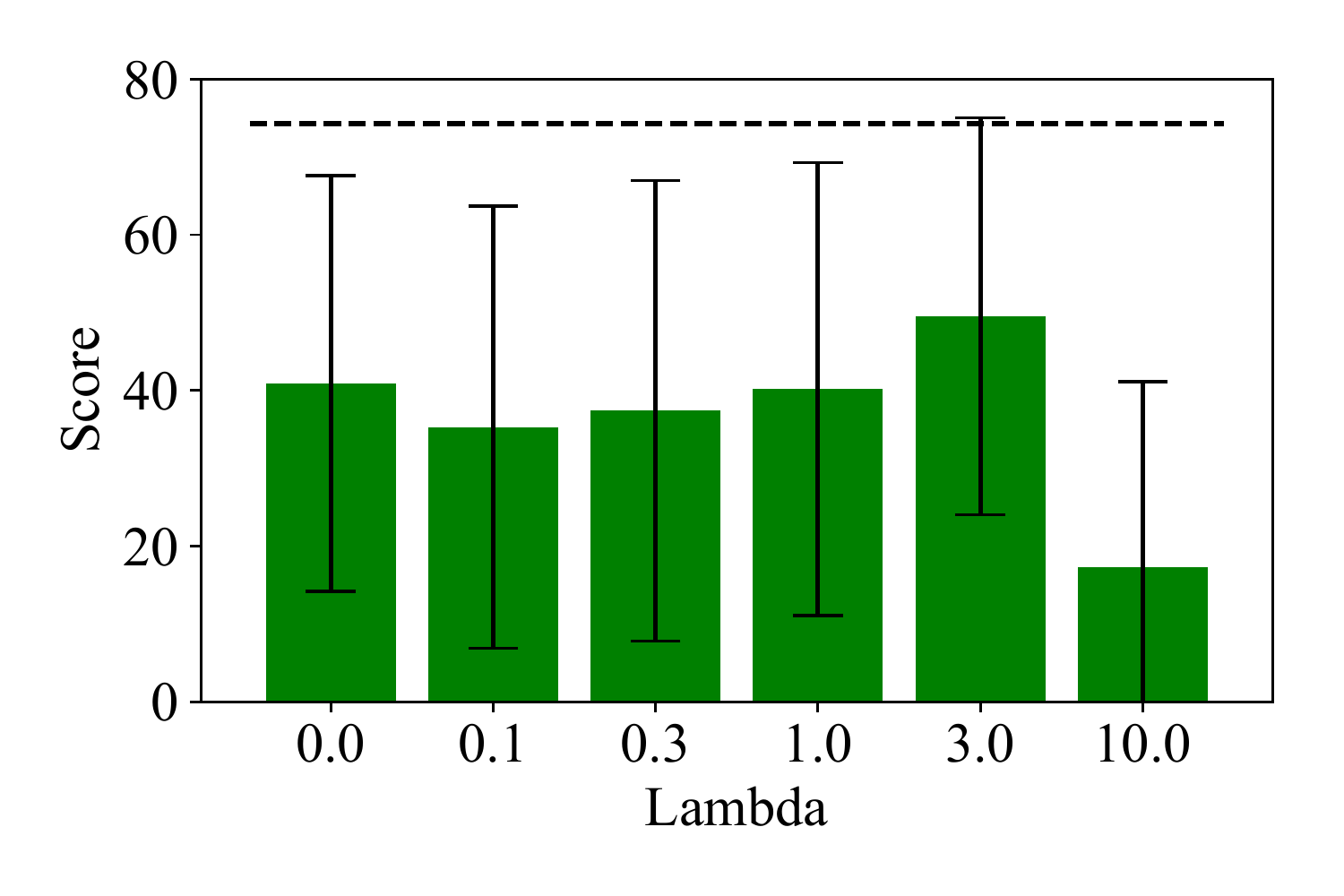}
  \end{minipage}
  \begin{minipage}{0.45\textwidth}
    \centering (b) Connector-Insertion
  \end{minipage}
  \caption{Comparison of MPC performance for different domain confusion loss coefficient $\lambda$ after 1,000 episodes. (a) and (b) show the performance for the Finger-Spin and Connector-Insertion task, respectively. The error bar indicates represents one standard deviation. The dashed lines show the average scores of the expert trajectories.}
  \label{fig:score_lambda}
\end{figure}

\subsection{Environments and hyperparameters} \label{sec:settings}
  We considered three tasks in MuJoCo physics simulator: Cup-Catch, Finger-Spin, and Connector-Insertion. Fig. \ref{fig:envs} shows the expert and agent domains for each task. For Finger-Spin, we make two different agent domains. One agent domain of Finger-Spin has different colors of objects and floors compared to the expert domain. The other agent domain of Finger-Spin has a different viewing angle in addition to the color shifts. It is difficult to train control policies by using only task rewards because all tasks provide only sparse rewards. Cup-Catch and Finger-Spin are instances of the DeepMind Control Suite~\cite{deepmindcontrolsuite2018}. We also built a new task, Connector-Insertion. In this task, an agent attempts to insert a connector to a socket. Constant rewards were obtained when the connector was in the socket. The position and angle of the connector and socket were initialized with random values at the start of the episodes. In this task, we added a constant bias to the action for moving the connector upward on the paper. This is equivalent to introducing domain knowledge that the socket exists upward on the paper.

  The contextual state and stochastic state sizes were 32 and 8 for all experiments. A small latent size is enough for DAC-SSM because domain-related information is eliminated from the latent space. The decoder refers to the domain labels to reconstruct domain-specific observation. The domain labels $y$ were simply concatenated to $h_t$ and $s_t$ as inputs to the domain conditional (DC) decoder. We used not only the DC decoder but also the DC encoder for the Finger-Spin of the tilted view. We implemented the DC encoder by training two separate encoders and switching them based on domain labels $y$. We use batches of 40 sequence chunks of 40 steps long for training. Except for the above mentioned, we adopted the same hyperparameters and architectures as PlaNet for the state space model. We implemented both the expert and domain discriminator as two layers of fully connected networks of 64 nodes with ReLU activation functions. The domain confusion loss coefficient $\lambda$ is 1.0 unless otherwise noted.
  
  We used CEM for planning algorithm. We adopted a short planning horizon $H=3$, number of optimization iterations $I=10$, number of candidate samples $J=4000$, and the best $K=20$ samples for refitting. The action repeats were 4, 2, and 800 for Cup-Catch, Finger-Spin, and Connector-Insertion, respectively. The action repeat for Connector-Insertion was extremely large because we set simulation timesteps of MuJoCo to a very small value of $5 \times 10^{-5}$; otherwise, objects easily pass through each other when they come into forceful contact. We evaluate three types of objectives for the planning: dual, imitation and task rewards. The dual rewards are weighted sum of task- and imitation-rewards with ratio of 10:1. The numbers of the expert and agent trajectories are both 100.

\subsection{Applying state representation to imitation learning with domain shifts}
  Fig. \ref{fig:dac-score} and Table \ref{tab:score} compares DAC-SSM using dual rewards (DAC/dual) to a baseline of existing SRL method (PlaNet/task) and naive implementation of the optimality discriminator with the baseline (PlaNet+$\mathcal{D}_\mathcal{O}$). DAC/dual achieved much higher performance for all tasks than the two baselines. This is because the domain-aware state representation of baselines does not help the agents to achieve higher performance via imitation learning with the domain shifts. We also compared DAC-SSM, a version using dual rewards (DAC/dual), a version using imitation rewards (DAC/imitation), and a version using task rewards (DAC/task). Except for Cup-Catch, DAC/imitation achieved the best performance. This is because the planning horizon length $H=3$ is too short for Finger-Spin and Connector-Insertion. We further trained our proposed model (DAC/dual) as well as versions with domain adversarial training but \textit{without} domain conditional encoders/decoders (DA/dual), and with domain conditional encoders/decoders but \textit{without} domain adversarial training (DC/dual). The performance of DAC/dual and DC/dual were almost the same, and that of DA/dual was much lower. In the settings of this experiment, the domain adversarial training was not effective because the domain confusion loss coefficient $\lambda=1$ was too small.
  
  Fig. \ref{fig:score_lambda} compares DAC/dual with $\lambda$ from 0.1 to 10.0 and DC/dual. DAC/dual with $\lambda=3$ achieved higher performance than DC/dual ($\lambda=0$) for Connector-Insertion. These results show that the obtained states on DAC-SSM help the agents to achieve the effective imitation learning with the domain shifts.

%\subsection{Imitation learning with plural expert domains}
%  We also considered the case in which the demonstration data were from plural domains. Fig. \ref{fig:plural_experts} shows examples of the expert data from plural domains of the Connector-Insertion task. Table. \ref{tab:score_plural} compares DAC/dual with $\lambda$ from 0.1 to 10.0 and DC/dual with plural expert domains. This result shows ...

\subsection{Reconstruction from State Representation}
  Fig. \ref{fig:reconstruction} shows the sequence of ground-truth examples and reconstructed images from the obtained state representation on DAC-SSM for Finger-Spin. The first 5 columns show context frames that were reconstructed from posterior samples, and the remaining images were generated from open-loop prior samples. The second and third row images were reconstructed from a sequence of states of $h_t$ and $s_t$ with domain labels $y$ via the DC decoder $p(o_t|h_t, s_t, y)$. Joint angles of the robotic arm and target object were successfully reconstructed from the states, whereas domain-dependent information (colors of the floor and object) depended on the domain labels. The last row images were reconstructed from the contextual states, $h_t$, \textit{without} domain labels $y$ using another decoder that was trained separately from our model. The joint angles were successfully reconstructed, whereas the colors appeared to be a mixture of the two domains. These results show that the obtained states via DAC-SSM have control-dependent information like the joint-angle, but do not have domain-dependent information like the colors which is not related to the control. In other words, we successfully acquire the domain-agnostic and task- and dynamics-aware sate representation via DAC-SSM.

\section{Conclusion and Discussions}
  We showed domain-agnostic and task- and dynamics-aware state representation was obtained via DAC-SSM. To obtain such state representation, we introduced domain adversarial training and domain conditional encoders/decoders into the recent task- and dynamics-aware sequential state space model. Moreover, we experimentally evaluated the MPC performance via imitation learning with the large domain shifts for continuous control sparse reward tasks in simulators. The state representation from DAC-SSM helped the agents to achieve comparable performance to the expert. The existing SRL failed to remove domain-dependent information from the states, and thus the agents could not perform the effective imitation learning with large domain shifts. We conclude that the domain-agnostic and control-aware states are essential for imitation learning with the large domain shifts, and such states are obtained via DAC-SSM.

  A question that remains is if DAC-SSM is applicable to larger and/or different types of domain shifts, e.g. modality-variant of data. Since the domain confusion loss coefficient $\lambda$ has task dependency as shown in Fig. \ref{fig:score_lambda}, we can expect better state representation is obtained by actively varying $\lambda$. Acquiring task-agnostic states to achieve a universal controller is also appealing future works. Imitation learning from human demonstration is challenging but interesting direction. This future work includes obtaining appropriate state representation from expert data \textit{without} action data. Implementation for real robotic tasks is another important direction for future works. Acquiring fully stochastic state representation is necessary for the real world tasks because the control system of the real robot have much larger uncertainty than simulation.

\section*{Acknowledgments}
  Most of the experiments were conducted in ABCI (\textit{AI Bridging Cloud Infrastructure}), built by the National Institute of Advanced Industrial Science and Technology, Japan.

\bibliography{iros2020}
\bibliographystyle{ieeetr}

\bigskip
\bigskip
\bigskip
\bigskip
\bigskip
\bigskip
\bigskip
\bigskip
\bigskip
\bigskip
\bigskip
\bigskip
\bigskip
\bigskip
\bigskip
\bigskip
\bigskip
\bigskip
\bigskip
\bigskip
\bigskip
\bigskip
\bigskip
\bigskip

\end{document}